\documentclass[letterpaper, 10 pt, conference]{ieeeconf}  

\IEEEoverridecommandlockouts                              

\usepackage{amsthm}
\usepackage{amsmath}
\usepackage{latexsym}
\usepackage{amssymb}
\usepackage{textcomp}
\usepackage{epsfig}
\let\labelindent\relax
\usepackage{enumitem}
\usepackage{dirtree}
\usepackage{listings}
\usepackage{float}
\usepackage[dvipsnames]{xcolor}
\usepackage[tight]{subfigure}
\usepackage{xcolor}
\usepackage{multirow}
\usepackage[noadjust]{cite}
\usepackage{hyperref}
\newcommand{\etal}{\textit{et al}.}
\hypersetup{
    colorlinks=true,
    linkcolor=blue,
    filecolor=magenta,      
    urlcolor=cyan,
    pdftitle={Overleaf Example},
    pdfpagemode=FullScreen,
    }
\usepackage{array}
\newcolumntype{P}[1]{>{\centering\arraybackslash}p{#1}}

\title{\LARGE \bf Learning to Detect Slip through Tactile Estimation of the Contact Force Field and its Entropy}
\author{Xiaohai Hu,$^{1}$ Aparajit Venkatesh,$^{1}$ Yusen Wan,$^{1}$ Guiliang Zheng,$^{1}$ Neel Jawale,$^{1}$ Navneet Kaur,$^{1}$\\ Xu Chen$^{1\ddagger}$ and Paul Birkmeyer$^{2}$ %
\thanks{$^{1}$Authors are with the Department of Mechanical Engineering, University of Washington. Email: {\tt\small{[huxh, venkat11, yusenwan, guiliang, neelj42, navneet, chx]}@uw.edu}. $^\ddagger$Corresponding author.}%
\thanks{$^{2}$Paul Birkmeyer is with Amazon Robotics. Email: 
 {\tt\small{paubirkm}@amazon.com}. This research was supported by a UW + Amazon Science Hub Gift-funded Robotic Research Project.}}%

\begin{document}

\maketitle
\pagestyle{empty}

\begin{abstract}
 Detection of slip during object grasping and manipulation plays a vital role in object handling. Existing solutions primarily rely on visual information to devise a strategy for grasping. However, for robotic systems to attain a level of proficiency comparable to humans, especially in consistently handling and manipulating unfamiliar objects, integrating artificial tactile sensing is increasingly essential. We introduce a novel physics-informed, data-driven approach to detect slip continuously in real time. We employ the GelSight Mini, an optical tactile sensor, attached to custom-designed grippers to gather tactile data.  Our work leverages the inhomogeneity of tactile sensor readings during slip events to develop distinctive features and formulates slip detection as a classification problem. To evaluate our approach, we test multiple data-driven models on 10 common objects under different loading conditions, textures, and materials. Our results show that the best classification algorithm achieves a high average accuracy of 95.61\%. We further illustrate the practical application of our research in dynamic robotic manipulation tasks, where our real-time slip detection and prevention algorithm is implemented.
\end{abstract}

\section{INTRODUCTION}
Tactile sensing, an essential sensory modality in humans, plays a pivotal role during object manipulation and grasping. It allows for the discernment of various object properties such as stiffness, weight, and surface texture. Remarkably, humans can adjust their grip force independently of visual feedback, highlighting tactile sensing's crucial role in this process. Johansson \etal, in their 1984 study \cite{roles}, demonstrated how humans utilize receptors in glabrous skin and sensorimotor memory for automatic precision grip control when handling objects with different surface textures. This research underscores a synergistic use of tactile feedback and sensorimotor memory in humans to predict and adjust to potential slip, thereby modulating grip force.

In the context of robotic systems, the incorporation of tactile-based slip detection plays a vital role in enhancing a robot's grip stability across varied operational dynamics.  Our study focuses on identifying key parameters for slip detection and developing a system for real-time slip detection that enables swift corrective actions. We explore innovative features for accurate slip identification and devise an automated system that can detect slips in real-time and implement corrective control strategies.
    \begin{figure}[h!]
    \centering
    \includegraphics[scale=0.4]{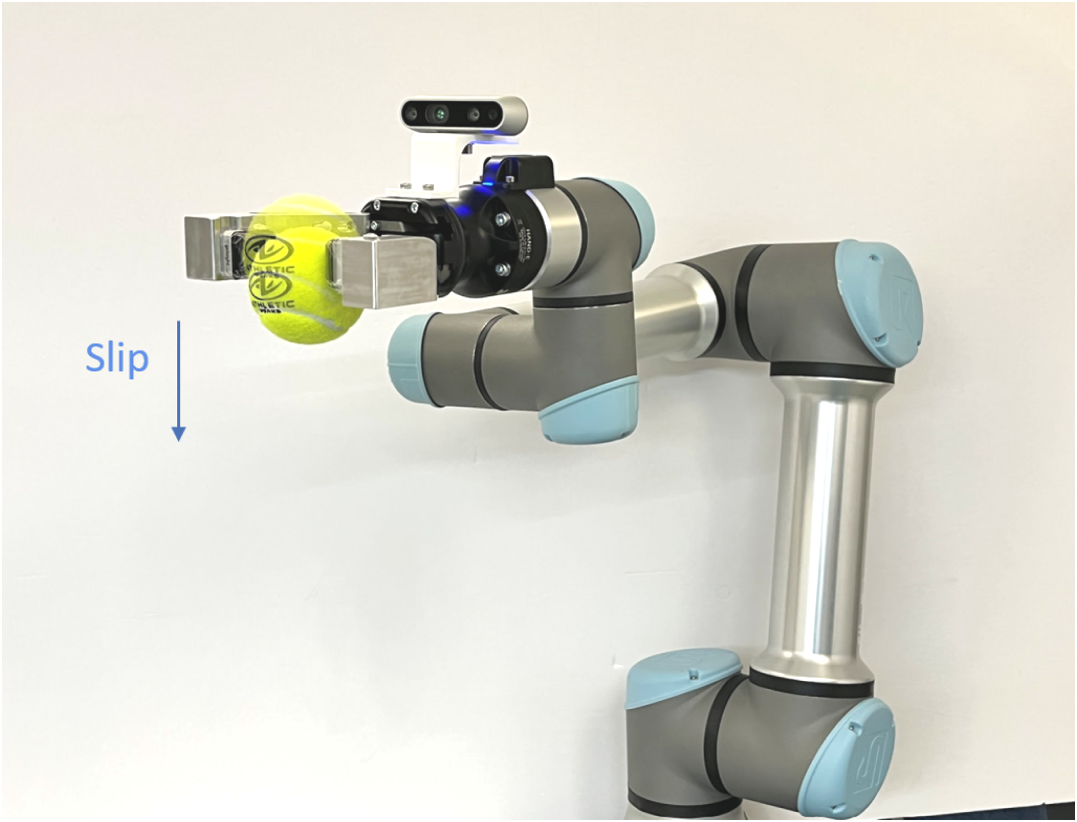}
    \caption{Experimental setup: An UR5e robot arm and a Robotiq parallel gripper, replacing the original fingertips with custom metallic adapters fitted with two GelSight tactile sensors. An Intel RealSense depth camera D435i is mounted atop the gripper.}
    \end{figure}

Our research introduces a novel approach to slip detection by harnessing the power of entropy analysis through GelSight sensor images, drawing inspiration from human sensory processing. This method centers on the extraction of entropy, a measure of randomness or disorder in the distributed marker displacements from the tactile images, to serve as a reliable indicator of potential slip events. By focusing on entropy, we can detect subtle changes in texture and surface roughness that are critical for slip detection, enhancing the system's effectiveness. This technique eliminates the need for prior knowledge about the object and grasping conditions, providing a substantial benefit in real-world scenarios where such information is often scarce or unavailable. The results of our study highlight that an entropy-focused strategy significantly boosts detection accuracy, proving to be more effective for slip detection and prevention in practical applications.

In the remainder of the paper, we will first introduce the basic working principle of the tactile sensor and related works in section II, followed by the hardware setup and the extraction of features from tactile images in section III. In section IV, we introduce how we collect the data and the classification methods we chose. We will then present the results given by various data-driven methods and validate the generalizability and future selection efficiency evaluation in section V. To demonstrate the real-world applicability of our proposed method, we illustrate its deployment in a robotic system that has been integrated with our slip detection and prevention algorithm, particularly emphasizing its performance during dynamic manipulation tasks.

\section{BACKGROUND AND RELATED WORK}

\subsection{Tactile Sensor}
The growth of tactile sensing is pivotal in robotic dexterity. 
Tactile sensing can be done intrinsically or extrinsically~\cite{xia}. 
Examples of intrinsic tactile sensing include measuring contact forces via joint torques~\cite{hand}\cite{opticaltorque} or transmission cable tension in equipped manipulators \cite{Jeong}. 
Extrinsic sensors, mounted on the robot hand’s exterior, use technologies like piezoelectric sensors for force and pressure detection~\cite{wettels}. Multi-modal sensors, such as BioTac~\cite{doe2014human}, provide varied data like force and temperature, but may lack resolution in complex force environments. Optical tactile sensors like Tactip~\cite{ward} and others~\cite{eth2022} provide representations of surface interactions for deducing force. 
Among these, the GelSight sensor stands out for its precise measurement of contact surface geometry~\cite{gelsight}~\cite{gelsight2}. 
The GelSight sensor 
typically utilizes a camera to record the deformation of a reflective, gel-coated transparent elastomer under applied force. This process generates images during surface contact, facilitating the creation of a depth map for the contact region.
Specifically, our project employs the GelSight Mini sensor, featuring marker dots on its cartridge for enhanced accuracy.
 \subsection{Slip Detection in Robotics}
Slip detection has been a focus in robotics for decades, with various methods proposed over the years. Starting with the skin acceleration sensor for slip and texture detection by Howe et al. in 1989~\cite{howe}, the field has evolved significantly. By 2004, methods like Ikeda et al.'s camera-based slip detection~\cite{ikeda}, and by 2012, Maldonado et al.'s fingertip sensing for object characteristics~\cite{Maldonado}, have been introduced.
Recent advancements primarily employ tactile sensing. Veiga et al. combined traditional tactile sensors with machine learning for 75\% accurate slip detection in 2015~\cite{Veiga}. Later developments include James et al.'s vision-based tactile sensors in 2018~\cite{james1}, Dong et al.'s incipient slip detection in 2018~\cite{incipient}, and Li et al.'s visual-tactile deep neural network in 2019~\cite{Li2018SlipDW}. Griffa et al.'s 2022 study used deep neural networks for force distribution classification~\cite{eth2022}, while Juddy et al. explored soft force sensors for deformable object grasping~\cite{soft}.

Particularly in tactile-based slip detection, Yuan et al.~\cite{wenzhen} discovered that the irregularity in gel deformation displacement relates well to slip. They quantified this by calculating the entropy of gel deformation length distribution in a histogram. Their experiments showed that entropy is an effective measure of partial slip. Notably, this entropy metric has not yet been widely used as a feature for slip classification in machine learning.

With the converging maturity of the tactile sensor on the one hand, and the evolution of robotic slip detection on the other, our research employs GelSight sensor images and entropy of the marker dots in the image to detect slip. By analyzing the randomness and its change of marker-dot displacement in these images, we can identify slips without needing prior knowledge about the object or grasping conditions. This advantage proves particularly valuable in real-world scenarios where such detailed information might be limited or unavailable.

\section{Hardware Setup And Tactile Data}

\subsection{Hardware Setup}
    \begin{figure}[h!]
    \centering
    \includegraphics[scale=0.45]{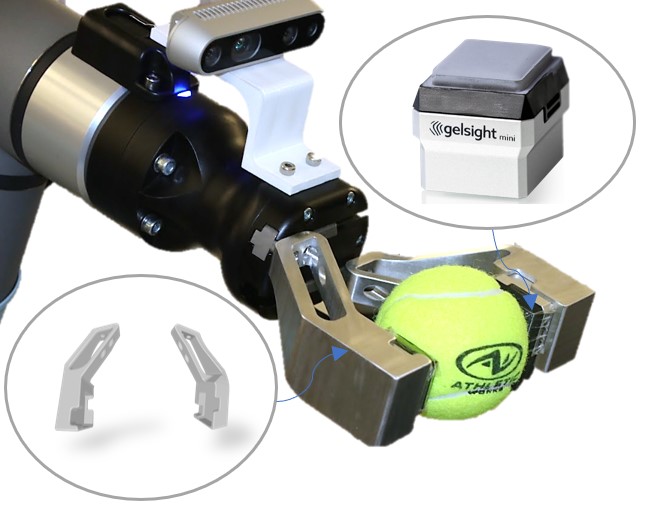}
    \caption{Design of adapter for housing Gelsight tactile sensors. A 45-degree flange extension was designed for the end-effector to extend the opening distance of the gripper to 85mm. The adapters are mounted on the Robotiq Hand-e adaptive gripper through four M2.5 screws.}
    \label{fig:hardware}
    \end{figure}
Our hardware setup includes a UR5e robotic arm, a Robotiq Hand-E gripper, and custom adapters for two Gelsight tactile sensors. The sensors are incorporated into tailored parallel end-effectors on the Robotiq Hand-e gripper and secured with screws. As shown in Fig. ~\ref{fig:hardware}, the end-effectors were designed to extend outwards along the movement directions of the gripper to create a large operational range of the system for a wide variety of target objects. The end-effector adapters, machined from 6061-T6 Aluminum for strength and stability, enable firm grasping of diverse objects.
The Gelsight mini sensor is capable of capturing high-resolution imprints of the contact surface, with a resolution of $3840\times 2160$. The sensor is coated with a Lambertian silicone gel layer and has a surface area of $18.6 (H) \times 14.3 (V)$ $mm^2$. Slip detection processes a single sensor reading. All equipment was operated using Ubuntu 20.04 on a PC equipped with an Intel Core i5-10210U CPU at a clock speed of 1.60GHz$\times$8. 
    \begin{figure}[htb!]
    \centering
    \includegraphics[scale=0.40]{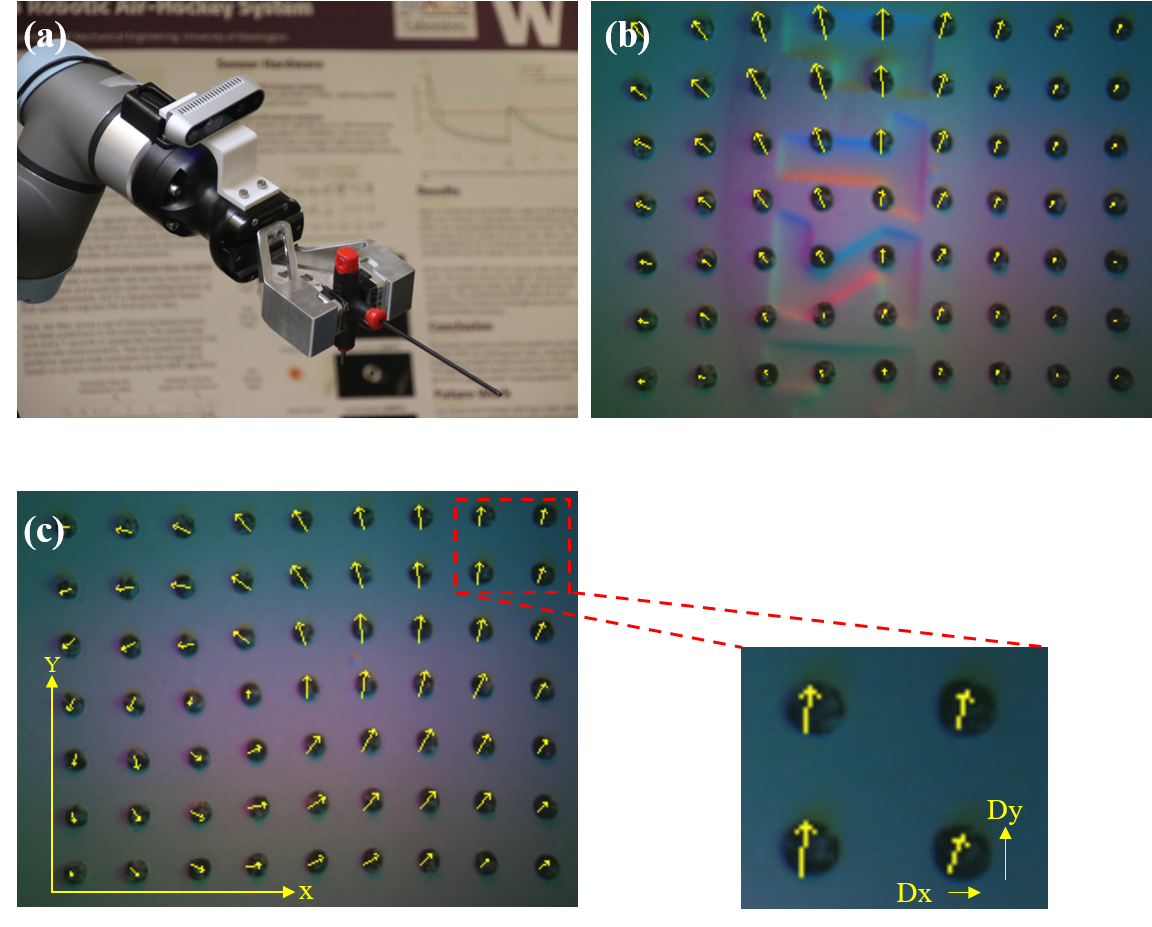}
\caption{\label{fig:grasping_slip}(a) Gripper grasping a T-handle hex key, (b) the displacement of individual markers overlaid on the tactile image,  and (c) zoomed-in section of GelSight image,  illustrates the gel deformation through arrows resulting from contact, denoted as $Dx_i$ and $Dy_i$, and referred to from hereon as the displacement field of the markers
}
\end{figure}
\subsection{Translating Tactile Frames to Features}
The camera within the GelSight tactile sensor captures the surface deformation dynamics as video sequences. This process is enhanced by the presence of 63 black markers, strategically arranged in a 7 $\times$ 9 matrix on the sensor's surface. These markers are crucial for enabling precise measurements of surface deformation. By observing the displacement of these markers over time, we can accurately determine the extent of deformation. We process each video frame and compare it to the initial frame to assess gel deformation. To quantitatively depict this deformation, yellow arrows are drawn on each frame (Fig.~\ref{fig:grasping_slip}). The specific methodologies employed are enumerated as follows:
We use the canny filter for marker selection and OpenCV's SimpleBlobDetector algorithm to determine their centers. By comparing marker positions from the initial to current frames, we compute each marker's average displacement~\cite{incipient} \cite{cable1}. 
In a stable grasp, the gel moves uniformly under the shear force indicated by evenly distributed arrows. However, during slip, this uniformity is disrupted, leading to increased entropy and inhomogeneity, as well as a significant rise in entropy derivative.  This increase in the derivative of entropy is considered a valuable feature for the classifier in subsequent stages. 
\\
\textbf{Feature $\bar{V_x}$ and $\bar{V_y}$}:
In addition to the raw displacement field, we observe that the rate of change of the vector field is also important to indicate slip.
To quantify these changes, we define the discrete-time mean velocity features $\bar{V_x}$ and $\bar{V_y}$ using the following equations:
\begin{equation}
\begin{aligned}
v_{x_i}(t) &= f\cdot[D_{x_i}(t) - D_{x_i}(t-\Delta t)]\\
\bar{V}_{x}(t) &= \frac{1}{n} \sum_{i=1}^{n} v_{x_i}(t) \\
\bar{V}_{y}(t) &= \frac{1}{n} \sum_{i=1}^{n} v_{y_i}(t) 
\end{aligned}
\end{equation}
Here, $f$(=25 Hz) is the sampling frequency, $n$(=63) represents the number of data points, $D_{x_i}$ and $D_{y_i}$ are the positions of the $i$-th data point at times $t$ and $t-\Delta t$, and $\Delta t$(=0.04s) is the sampling interval.

\begin{figure}[h!]
    \centering
    \includegraphics[scale=0.6]{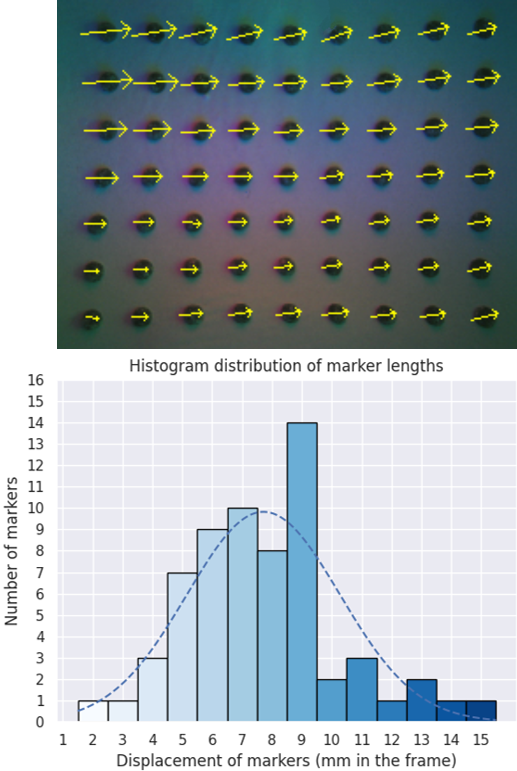}
    \caption{This histogram illustrates the distribution of marker flow, depicting the frequency of various marker lengths. Each bar in the histogram represents a distinct state, with its height indicating the relative occurrence of that state within the overall distribution.}
    \label{fig:entrop_histo}
\end{figure}

\textbf{Feature $E$ and $\frac{\delta E}{\delta t}$}: The marker flow displacement inhomogeneity, quantified as entropy, serves as a key metric in our slip detection approach. We define this entropy, \(E(X)\), mathematically as:
\begin{equation}
E(X) = - \int_{X} p(x) \log p(x) dx
\end{equation}
While there are various entropy metrics available, our analysis uses the discrete version of Shannon entropy. This choice is underpinned by its validation in prior research, as exemplified in Yuan et al. \cite{wenzhen}. In this context, the histogram \(X\) represents the frequency distribution of the magnitude of the displacement field, and \(p(x)\) is the corresponding probability density function. Entropy calculation involves multiplying each histogram value by its probability and then applying the negative logarithm to this product.
 For example, Fig. \ref{fig:entrop_histo} shows a tactile image with 63 arrows of varying lengths and the corresponding histogram. 
The proposed entropy calculation involves summing over the discrete values of 
 \(X\), which consists of 15 distinct values,  \(\{x_1, x_2, \ldots, x_{15}\}\) and their associated probabilities \(\{p(x_1), p(x_2), \ldots, p(x_{15})\}\), based on:

\begin{equation}
E(X) = - \sum_{i=1}^{15} p(x_i) \log p(x_i)
\end{equation}

\begin{figure}[h!]
    \centering
    \includegraphics[scale=0.25]{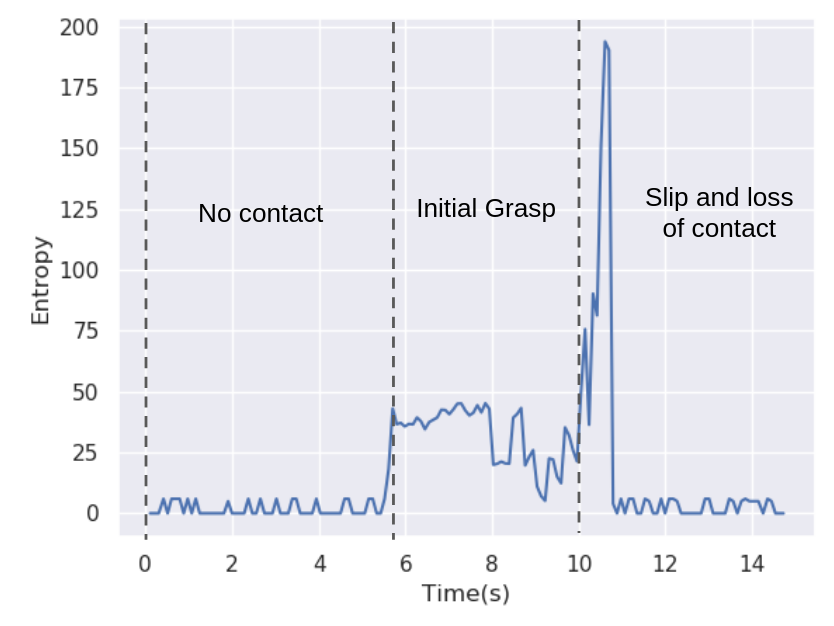}
    \caption {\label{fig:entropy_process}A slip trial was conducted to illustrate the change of entropy from the no contact to object, through the initial grasp, to the incipient slip, and ultimately to the loss of contact to object. Initially, a rectangular cardboard box was held between the grippers (embedded with GelSight sensor) of the robot, at this stage the entropy was almost zero. The grippers were brought closer together till there was an initial contact and a gentle grasp of the box, leading to a notable increase and subsequent stabilization of entropy. Following this, the robotic arm was maneuvered in a manner that induced slippage of the object. At the moment of slip, a sharp spike in entropy was observed and then the entropy returned zero once the robotic gripper with the tactile sensor lost complete contact.}
    \end{figure}

When an object begins to slip, the displacement field becomes more inhomogenous due to the non-uniform contact forces that arise throughout the contact surface as the object moves. The inhomogeneity is more significant around the edges of the contact region, resulting in a non-uniform displacement field and increased entropy, as illustrated in Fig. \ref{fig:entropy_process}.

To see another major design of this paper, consider the entropy changes during a grasp in Fig. \ref{fig:entropy_process}. We see that the entropy increases when the grippers initially come into contact with the object, and this value remains relatively constant as long as the object is securely grasped. However, when the object begins to slip, a sharp increase in entropy is observed. Thus, it is evident that reasonably high entropy values can exist even when an object is securely grasped. Additionally, the entropy remains almost constant when a secure grasp is established. To further enhance the classification of slip, we propose integrating the rate of change of entropy as another feature to feed into the classifier. The rate of change of entropy is calculated as follows:
\begin{equation}
   \frac{\delta E}{\delta t}\approx f\cdot [E(t)-E(t-\Delta t)]
\end{equation}
where $E$ is the entropy and $\Delta t$ is the sampling time. Since entropy varies with object characteristics, we trained a classifier using data from objects of diverse shapes, sizes, and materials to categorize slips effectively.

\section{Method}
\subsection{Data Acquisition}
\begin{figure}[h!]
    \centering
    \includegraphics[width=0.45\textwidth]{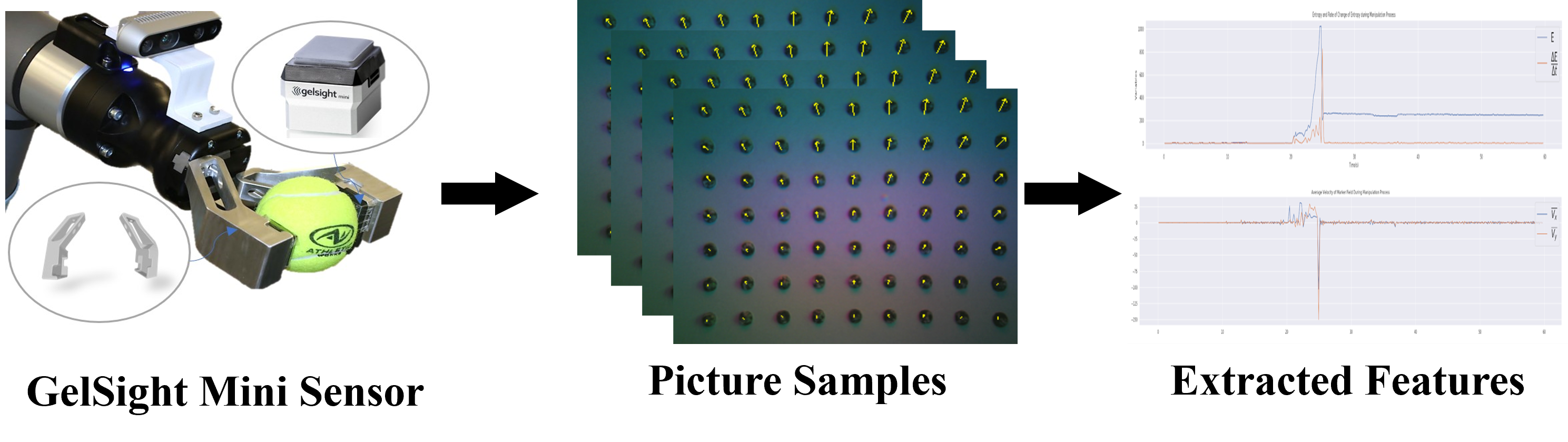}
    \caption{\label{fig:data acquisition}Illustration of the data acquisition process. Initially, the manipulator grips the object and initiates movement. Subsequently, the sensor captures the grasping details and proceeds to extract features from these samples.}
\end{figure}

 To collect the necessary data, we selected 10 objects of different materials and shapes that are commonly encountered. The process is shown in Fig. \ref{fig:data acquisition}.

A predetermined grasping sequence was meticulously devised, incorporating a range of motion primitives. These included translational and rotational movements across all axes of the Cartesian coordinate system, as well as planar and spherical motions, culminating in twelve distinct motion primitives.
In our study, we emphasize the detection of object slippage during manipulation tasks. To enhance the generalizability of our findings, we placed test objects at various locations on a table. The manipulator was programmed to approach the object, grasp it, lift it to a standardized position, and then execute a pre-defined grasping sequence during which we continuously collected tactile sensor data.
For each grasping trial, an external load was incrementally applied by human operators. A parallel jaw gripper was deployed using position control to ensure that the object was grasped with proper tightness. In tests aiming to avoid slip (static cases), the applied load or rotation was intentionally limited to maintain a low force threshold, guaranteeing the absence of slippage. Conversely, for the slip dataset, objects were either manually nudged or rotated with care. This was done to induce both translational and rotational slippage without exerting excessive force that could potentially harm the tactile sensor's gel surface.

For each frame captured by the tactile sensor, we identified whether it represented slippage or not. Nevertheless, there were instances where the object transitioned between static and slip states. In such cases, the data labeling was rectified based on human judgment. For example, if an object's state evolved from static (non-slip) to slip during the sequence, the exact frame marking this transition was documented. Frames preceding this transition were classified as non-slip, while subsequent frames were labeled as slip. Our labeling methodology is congruent with the approaches delineated in the works of ~\cite{eth2022} and ~\cite{incipient}.
\begin{figure*}[h!]
  \centering
  \includegraphics[width=\textwidth]{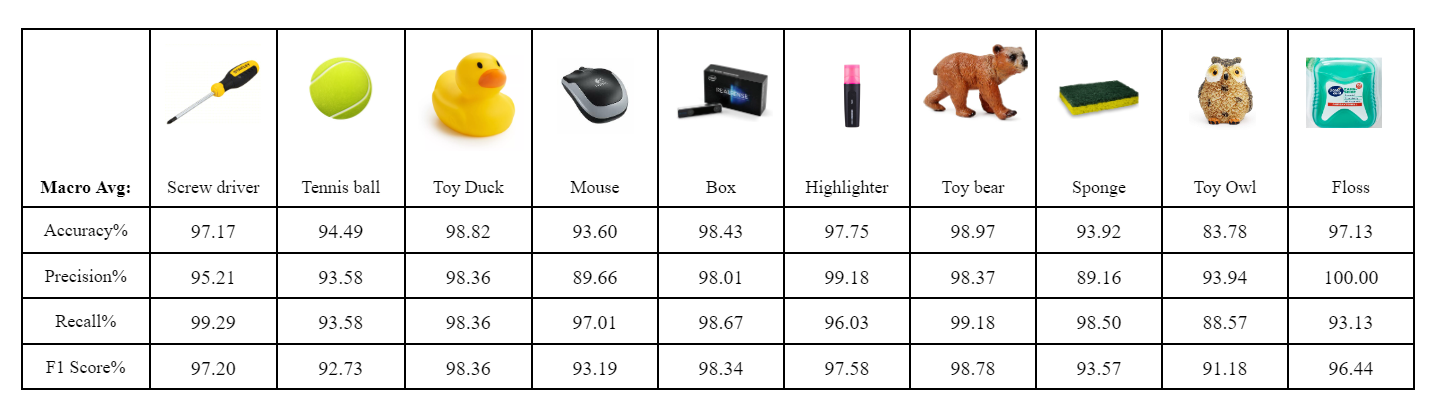}
  \caption{\label{fig: performance}Performance of the developed slip detection for objects with different materials and surface characteristics. }
\end{figure*}

\subsection{Classification Method}
Following the feature extraction phase, we utilized machine learning classifiers to identify the states of object grasping. Given the inherent nonlinearity in the task, we experimented with various classifiers to understand the intricate slip dynamics. We selected
four algorithms:

\begin{itemize}
\setlength\itemsep{0em}
\item Support Vector Machine (SVM): $g(x) = \text{sign}(\mathbf{w}^T\mathbf{x} + b)$
\item Random Forest (RF): $g(x) = \frac{1}{T}\sum_{t=1}^T h_t(x)$
\item K-Nearest Neighbour (KNN): $g(x) = \text{mode}(y_i : x_i \in N_k(x))$
\item Decision Tree (DT): $g(D,A) = H(D) - H(D|A)$
\end{itemize}

In this context, $g(x)$ indicates the likelihood that the input $x$ is associated with a specific class. The terms $\mathbf{w}$ and $b$ represent the weight vector and bias in SVM,  while $h_t(x)$ denotes the prediction by the $t$-th tree in RF. KNN uses $\text{mode}$ to find the most frequent class among $x$'s $k$ nearest neighbors. The DT $g(D,A)$ calculates information gain, assessing feature importance. In the next section, we will assess these classifiers' accuracy using our extracted features.

\section{Experiment Results}
Experiments are performed to analyze the proposed new features. To thoroughly gauge their efficacy, we juxtapose these features against basic alternatives such as $\bar{V_x}$ and $\bar{V_y}$. These experiments consist of two parts: Part I focuses on classification accuracy, and Part II sheds light on generalization capabilities. Additionally, we visualize the effectiveness of feature selection, examine the computational time of inference, and provide a demonstration of using the proposed method to slide a book out of a shelf while preventing slip. 

\subsection{Experiment Sets}
For these experiments, we implemented the chosen classifiers via Python's scikit-learn
library. Our dataset includes approximately 14,000 samples among ten different objects. Here, a "sample" refers to a sequence of tactile images extracted from the sensor's video stream, from which we derive features for classifier training. And the features of one sample consist of one feature vector, which is the input of ML model. Thus, one sample has one feature vector and there are 14,000 feature vector in the dataset. To assess the accuracy and generalization of the proposed features, we conducted two sets of experiments. In the first, we utilized the four machine learning classifiers mentioned earlier to assess the accuracy of our features.  By consolidating all samples, we adopted a five-fold cross-validation technique. This method splits the samples into five sections, each holding 2,800 samples. During each iteration, four sections were designated for classifier training, leaving one for evaluation. The final accuracy was the average over the five cycles. In the second experiment, we focus on KNN, the best-performing classifier from the first experiment, to gauge the feature's adaptability across various objects. In this case, we divided the samples based on object categories, resulting in ten distinct subsets of samples. Here, samples were categorized based on object types, creating ten unique sample subsets. In each test, nine subsets were used for training and the tenth for evaluation. It is important to highlight that the classifiers were tasked with categorizing samples from unfamiliar objects not present in the training dataset. The accuracy thus reflects the feature's adaptability.
In our experiments, we extracted four key features: $\bar{V_x}$, $\bar{V_y}$, $E$, and $\frac{\delta E}{\delta t}$. Classifiers were trained using three feature sets: $\bar{V_x}$ and $\bar{V_y}$ only, solely $E$, and $\frac{\delta E}{\delta t}$, and a combination of all four features. To determine feature effectiveness, we compared results across these experimental setups. Accuracy assessment involved four metrics: accuracy, recall, precision, and F1 score. ``Accuracy'' signifies the percentage of correctly classified instances relative to the total number of instances. ``Precision'' measures the classifier's ability to distinguish true positives from false positives, with both precision and accuracy ideally being high for a perfect classifier. ``Recall'' assesses the classifier's capacity to distinguish true positives from false negatives. The F1 score, which is the harmonic mean of precision and recall, offers a balanced assessment of model performance.

For hyperparameter tuning, a grid search was conducted during training for each classification algorithm. The best results were observed with KNN classifier, setting the nearest neighbor hyperparameter set to 1, and SVM using the RBF kernel and a regularization parameter of 1. Hyperparameter adjustments did not enhance the performance of the RF and DT classifiers, so their default settings were maintained.

\subsection{Experiment I Result: Accuracy of Slip Detection}
The outcomes of Experiment I, as detailed in Table \ref{table:cross validation}, reveal that incorporating entropy and its rate of change significantly enhances classifier accuracy across all models. Replacing $\bar{V_x}$ and $\bar{V_y}$ with  $E$, and $\frac{\delta E}{\delta t}$ resulted in substantial accuracy improvements for DT, RF, SVM, and KNN by 11.7\%, 6.7\%, 4.9\%, and 11.2\%, respectively. Other measures like Precision, Recall, and F1 score also saw improvements, with SVM being the top performer in this setup. Utilizing all four features combined, compared to using only $\bar{V_x}$ and $\bar{V_y}$, led to notable accuracy increases of 9.92\%, 8.36\%, 5.58\%, and 15.37\% for DT, RF, SVM, and KNN, respectively. In this scenario, KNN demonstrated superior performance. Thus,  $E$ and $\frac{\delta E}{\delta t}$ outperform $\bar{V_x}$ and $\bar{V_y}$ in terms of accuracy, enhancing overall model performance.

Fig. \ref{fig: performance} showcases the classification accuracy of different objects using KNN. Most objects exhibit an accuracy greater than 95\%, with only the owl falling below 90\%. The results show that the performance of the classifier for each category is balanced, and the proposed features have broader usage rather than can be used on specific objects.

\begin{table}

\centering
\caption{\label{table:cross validation} 5 fold cross-validation result of Selecting Different Features for Training.}
\begin{tabular}{cc|cccc}
\hline
\textbf{Experiment Set} & \textbf{Metrics} & \textbf{DT} & \textbf{RF} & \textbf{SVM} & \textbf{KNN} \\
\hline
\multirow{4}{*}{$\bar{V_x}$ and $\bar{V_y}$} 
& Accuracy & 82.20 & 84.22 & 89.30 & 80.25 \\
& Precision & 94.72 & 92.63 & 83.30 & 94.25 \\
& Recall & 75.91 & 79.56 & 93.58 & 74.66 \\
& F1 & 83.94 & 85.15 & 88.13 & 82.68 \\
\hline
\multirow{4}{*}{$E$ and $\frac{\delta E}{\delta t}$} 
& Accuracy & 93.88 & 90.99 & 94.22 & 92.45 \\
& Precision & 90.52 & 86.60 & 88.60 & 90.80 \\
& Recall & 96.47 & 93.87 & 99.20 & 93.43 \\
& F1 & 93.38 & 90.08 & 93.60 & 92.03 \\
\hline
\multirow{4}{*}{All features}
& Accuracy & 92.07 & 92.59 & 94.88 & 95.61 \\
& Precision & 96.73 & 95.21 & 91.19 & 95.07 \\
& Recall & 89.09 & 90.55 & 97.94 & 95.76 \\
& F1 & 92.49 & 92.59 & 94.44 & 95.40 \\
\hline
\end{tabular}
\end{table}

\begin{table*}[tb]

\centering
\caption{\label{table: object validation}Cross-validation results for 10 different objects.}
\begin{tabular}{c|c|cccccccccc}
\hline
\multirow{2}{*}{Feature Set} & \multirow{2}{*}{Metric} & \multicolumn{10}{c}{Experiment Index and Valid Object} \\
& & 1 & 2 & 3 & 4  & 5  & 6  & 7 & 8  & 9  & 10  \\ &&ball & bear &  box & duck & floss &  highlighter & mouse & owl & screwdriver & sponge\\
\hline
\multirow{4}{*}{$\bar{V_x}$ and $\bar{V_y}$} 
& Accuracy \%  & 42.92 & 47.93 & 45.95 & 48.79 & 40.13 & 49.29 & 42.69 & 79.27 & 53.5 & 38.46 \\
& Precision \% & 89.81 & 88.82 & 75.38 & 90.1  & 91.78 & 86.5 & 89.42 & 77.43 & 89.42 & 88.2 \\
& Recall \% & 45.12 & 44.76 & 44.04 & 42.87 & 40.94 & 48.29 & 44.96 & 100.00 & 53.91 & 40.54 \\
& F1 \%  & 60.06 & 59.53 & 55.6 & 58.1 & 56.62 & 61.98 & 59.84 & 87.28 & 67.26 & 55.55 \\
\hline
\multirow{4}{*}{All Features}
& Accuracy \% & 71.24 & 80.72 & 77.84 & 89.07 & 61.81 & 84.02 & 70.44 & 91.06  & 58.37 & 84.63 \\
& Precision \% & 93.52 & 94.57 & 92.31 & 96.25 & 91.17 & 93.09 & 90.41 & 90.27 & 96.96 & 91.15 \\
& Recall \% & 63.52 & 70.64 & 68.89 & 80.05 & 53.00 & 77.82 & 63.34 & 100.00 & 56.43 & 77.54 \\
& F1 \% & 75.66 & 80.87 & 78.90 & 87.41 & 67.03 & 84.77 & 74.49 & 94.88 & 71.34 & 83.80 \\
\hline
\multirow{4}{*}{$E$ and $\frac{\delta E}{\delta t}$}
& Accuracy \%  & 91.67 & 86.91 & 91.09 & 97.58 & 93.16 & 89.27 & 87.21 & 91.73  & 97.46 & 90.66 \\
& Precision \% & 90.43 & 88.18 & 90.35 & 94.45 & 92.15 & 91.99 & 91.26 & 91.00 & 95.90 & 89.60 \\
& Recall \% & 91.99 & 82.63 & 89.85 & 99.37 & 91.81 & 86.43 & 83.48 & 100.00 & 99.32 & 89.04 \\
& F1 \% & 91.20 & 85.32 & 90.10 & 96.85 & 91.98 & 89.13 & 87.20 & 95.29 & 97.58 & 89.32 \\ 
\hline
\end{tabular}
\end{table*}

\subsection{Experiment II Result: Generalizability of the Method}
The results of Experiment II, presented in Table \ref{table: object validation}, involved training classifiers with samples from nine objects and testing them with samples from a different, single object each time. In the table, the object name indicates the test object. For instance, ``box'' implies training with samples from all objects except the box and testing with box samples.

The findings suggest classifiers perform better with entropy ($E$) and its rate of change ($\frac{\delta E}{\delta t}$) than with $\bar{V_x}$ and $\bar{V_y}$ when encountering unfamiliar objects. A comparison of Tables \ref{table:cross validation} and \ref{table: object validation} shows a decline in accuracy when facing new objects. However,when using $E$ and $\frac{\delta E}{\delta t}$ as features, the accuracy only decrease less than 6\% compared with the accuracies in experiment I, which is significantly less. In contrast, reliance on $\bar{V_x}$ and $\bar{V_y}$ leads to a more substantial decrease in accuracy; for example, accuracy for ``sponge'' drops to 38.46\% and ``mouse'' to 42.69\%. The decline is also notable when all features are used together.

These results lead to the conclusion that $E$ and $\frac{\delta E}{\delta t}$ offer better generalization capabilities. They can be directly applied to new objects without substantial performance reduction. Conversely, the generalizability of $\bar{V_x}$ and $\bar{V_y}$ is limited, and they can negatively impact classification. The use of all features combined does not enhance model performance compared to using only $E$ and $\frac{\delta E}{\delta t}$. Additionally, we assessed the performance of the trained classifier on unseen objects, as shown in Table \ref{table: unseen objects}. The results indicate that the classifier maintains good accuracy with unfamiliar objects.
\begin{table}[tb]
\centering

\caption{\label{table: unseen objects}Performance Metrics for Unseen Objects.}
\begin{tabular}{c|cccc}
\hline
Unseen Object   & Accuracy\% & Precision\% & Recall\% & F1\%  \\
\hline
Book            & 71.69      & 88.45       & 59.17    & 70.91 \\
Raccoon         & 93.73      & 97.03       & 91.02    & 93.93 \\
Contact Solution & 86.66    & 95.09       & 80.32    & 87.08 \\ 
\hline
\end{tabular}
\end{table}

\subsection{Feature Visualization and Analysis}

\begin{figure}[tb]
\centering
\includegraphics[width=0.45\textwidth]{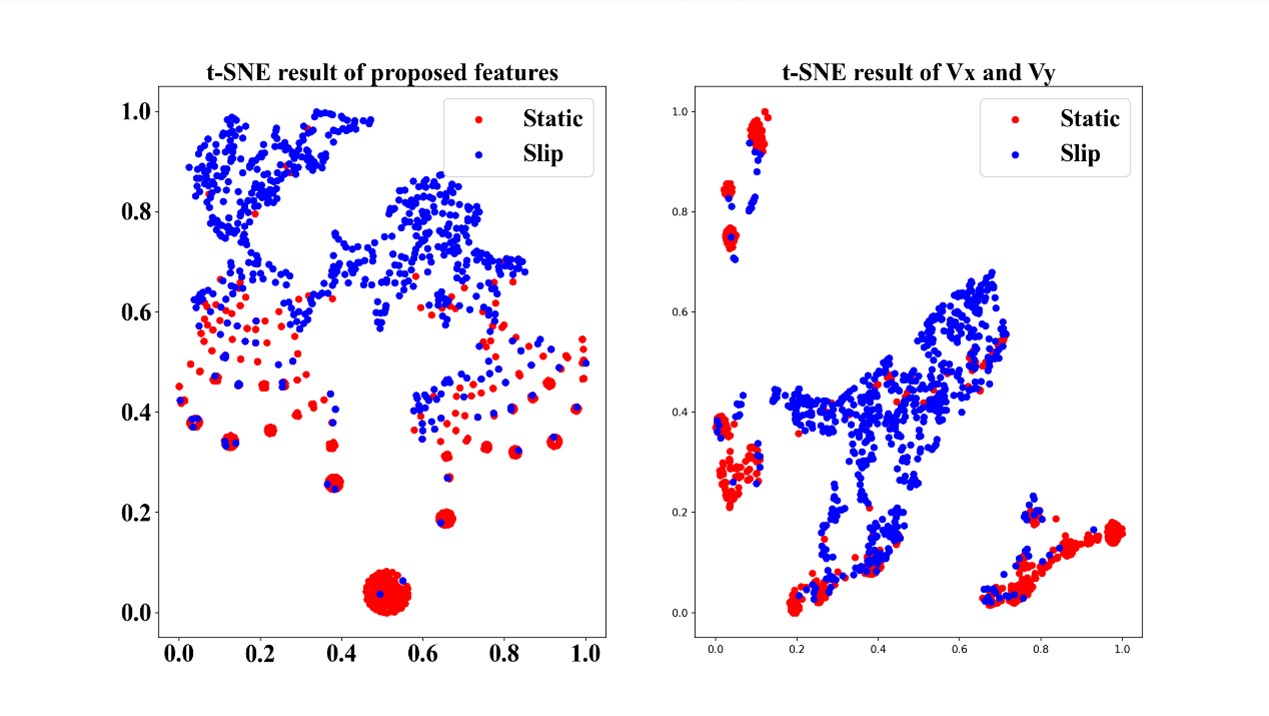}
\caption{\label{fig: visualization}t-SNE visualization of features: (Left) Entropy ($E$) and its rate of change ($\frac{\delta E}{\delta t}$); (Right) $\bar{V_x}$ and $\bar{V_y}$.}
\end{figure}

As shown in Fig. \ref{fig: visualization}, we employ t-SNE to visualize the extracted features, where each point represents a sample characterized by its respective features. In Fig. \ref{fig: visualization}(a),  the division between $E$ and $\frac{\delta E}{\delta t}$ is clearly evident. Despite some overlap, samples from different classes predominantly occupy distinct areas. Conversely, Fig. \ref{fig: visualization}(b) illustrates that the visualization of $\bar{V_x}$ and $\bar{V_y}$ is less distinct. The points representing ``static'' and ``slip'' classes significantly overlap, indicating less effective separation. This visualization suggests that $E$ and $\frac{\delta E}{\delta t}$ lead to more pronounced classification behavior. 

\begin{figure*}[tb]
     \centering
  \includegraphics[width=0.95\textwidth]{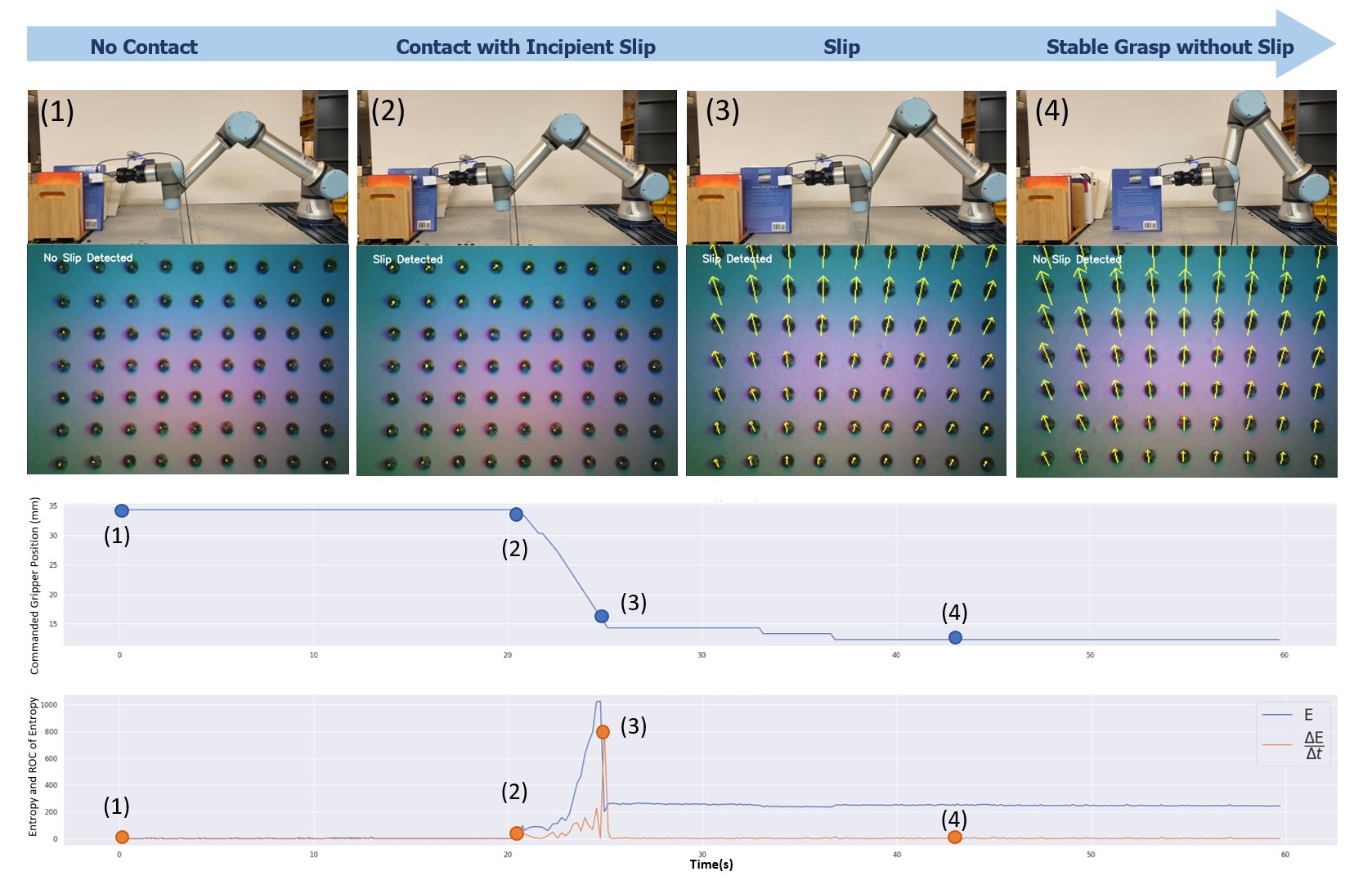}
  \caption{\label{fig:bookslidingdemo}A demonstration of sliding a book out of a shelf is presented, consisting of multiple stages of grasping. The initial row of images portrays the progressive grasping stages, commencing from (1) a static initial grasp, advancing towards (2) an incipient slip at the start of manipulation, further to (3) an actual slip, and culminating at (4) a stable grasp. The subsequent row exhibits the data obtained from a tactile sensor and the corresponding real-time detection of slip. The third row displays the command gripper distance to forestall slippage, whereas the fourth row depicts $E$ and $\frac{\delta E}{\delta t}$ throughout the grasping procedure. } 
\end{figure*}

\subsection{Computational Time of Inference}
During each of the 10 trials, we captured 200 prediction values and evaluated the classifiers' inference times using Python's \texttt{time.time()}. With the GelSight mini tactile sensor operating at 25 FPS, there is a 40 ms window for each frame processing. The average inference times and their usage percentages are:
\begin{itemize}
    \item Support Vector Machine (SVM): 0.33 ms (0.825\% of frame time)
    \item Random Forest (RF): 0.54 ms (1.35\% of frame time)
    \item K-Nearest Neighbour (KNN): 0.94 ms (2.35\% of frame time)
    \item  Decision Tree (DT): 0.29 ms (0.725\% of frame time)
\end{itemize}
All classifiers efficiently fit within the available frame processing time, making them suitable for real-time tactile image processing with the GelSight mini sensor.

\subsection{Sliding Out a Book from a Shelf}
 In this experiment, a UR5e robot performed a task to retrieve and reposition a book from a shelf.  Lacking prior knowledge about the book's weight and stiffness, we adopted a cautious grasping approach with minimal clamping force to avoid potential damage. The task began with positioning the parallel gripper adjacent to the bookshelf without making initial contact with the book.  The gripper was then maneuvered towards the book, and a safe grasp was executed using appropriate parameters, which entailed a relatively low grasping force. No slip occurred during the pre-manipulation phase. However, during the subsequent stage of extracting the book, the absence of a slip detection method and corresponding prevention led to slippage, hindering task completion.

Repeating the experiment with our slip detection algorithm and a slip-prevention force control integrated into the robotic system showed significant improvement. The algorithm promptly detected the initial slip during book extraction. Despite the gradual increase in grasping force, the smooth book cover and the drag exerted by the gripper caused the incipient slip to increase. The slip-prevention algorithm responded by gradually increasing the grasping force until a stable grasp was achieved, which was maintained for a brief duration to ensure secure gripping of the book. While maintaining the grasping force, the robot slid the book out of the bookshelf and adjusted the force as necessary to prevent further slips. After the book is extracted, the grasping force is gradually reduced to release the book. With the slip detection and prevention algorithms implemented, the grasp was continuously sustained, and the manipulation task was accomplished using the same set of initial grasping parameters.

Fig. \ref{fig:bookslidingdemo} shows the grasping stages of the book retrieval, along with corresponding $E$ and $\frac{\delta E}{\delta t}$. The figure indicates that $E$ and $\frac{\delta E}{\delta t}$ were almost negligible when the book is held with safe grasping parameters. However, as the manipulation process begins,  $E$ and $\frac{\delta E}{\delta t}$ slightly increased, indicating the need for a grip adjustment. As the manipulation process proceeds, those values rose, indicating the occurrence of more slips. To prevent slip, the gripping pose was further modified until the $E$ reached a constant value and $\frac{\delta E}{\delta t}$ approached zero. It is noteworthy that the $E$  value is higher at the end of the manipulation process than at the start, which can be attributed to the dynamic forces acting on the book during the manipulation process, resulting in non-homogeneity in the marker field. 

The classifier successfully attained a 71.69\% accuracy in categorizing the book's grasping state. This experiment illustrates that effectively monitoring entropy changes serves as a reliable guide for slip detection and prevention, thereby enhancing the success of manipulation tasks. 

\section{CONCLUSIONS AND FUTURE WORK}
In this paper, we proposed a novel approach for continuous slip detection using modern optical tactile sensors. Our method employs a physics-informed, data-driven strategy that leverages the distributed contact force field, its entropy, and the rate of change of entropy extracted from tactile sensors.
 By reliably monitoring incipient slip, our approach predicts slip occurrences during object grasping tasks. Additionally, once the slip detection classifier is trained on a sufficient number of objects across different classes, the method has the generalizability to be applied for slip detection and prevention on previously untrained objects.
 
Further work includes developing algorithms to control the slip of objects while performing manipulation and grasping tasks.
Parameters that are good indicators of slip have been identified in this paper, and work can be done to determine how these parameters can be utilized to control slip. 

\addtolength{\textheight}{0cm}   




\section*{ACKNOWLEDGMENT}

The authors would like to thank Siyuan Dong and Yu She for their expertise in using the sensor. Thanks to Hui Xiao for suggestions and discussion. Our appreciation also goes to  Michael Wolf for his coordination and feedback along the way.
This work was supported by a UW+Amazon Science Hub Gift-funded Robotic Research Project.

\bibliographystyle{IEEEtran}
\bibliography{mycite}
\end{document}